\documentclass[sigconf]{acmart}

\def\BibTeX{{\rm B\kern-.05em{\sc i\kern-.025em b}\kern-.08emT\kern-.1667em\lower.7ex\hbox{E}\kern-.125emX}}
    
\copyrightyear{2019}
\acmYear{2019}
\setcopyright{acmlicensed}
\acmConference[HRI '19 Workshop]{HRI '19: ACM Workshop on Social Human-Robot Interaction of Human-care Service Robots}{March 11--14, 2019}{Daegu, Korea}
\acmBooktitle{HRI '19: ACM Workshop on Social Human-Robot Interaction of Human-care Service Robots, March 11--14, 2019, Daegu, Korea}
\acmPrice{15.00}
\acmDOI{10.1145/1122445.1122456}
\acmISBN{978-1-4503-9999-9/18/06}

\usepackage{graphicx}
\usepackage[lofdepth,lotdepth]{subfig}

\begin{document}

%
% The "title" command has an optional parameter, allowing the author to define a "short title" to be used in page headers.
\title{Enabling Socially Competent navigation \\through incorporating HRI}

%
% The "author" command and its associated commands are used to define the authors and their affiliations.
% Of note is the shared affiliation of the first two authors, and the "authornote" and "authornotemark" commands
% used to denote shared contribution to the research.>
\author{Arturo Cruz-Maya}
\email{arturo.cruzmaya@softbankrobotics.com }
\orcid{ 0000-0003-3526-0307}
\author{Fernando Garcia}
\email{ferran.garcia@softbankrobotics.com }
\author{Amit Kumar Pandey}
\email{akpandey@softbankrobotics.com }
\affiliation{%
  \institution{Softbank Robotics Europe}
  \streetaddress{43 Rue du Colonel Pierre Avia}
  \city{Paris}
  \state{Ile de France}
  \postcode{75015}
}

%
% By default, the full list of authors will be used in the page headers. Often, this list is too long, and will overlap
% other information printed in the page headers. This command allows the author to define a more concise list
% of authors' names for this purpose.
\renewcommand{\shortauthors}{Cruz-Maya, et al.}

%
% The abstract is a short summary of the work to be presented in the article.
\begin{abstract}
Over the last years, social robots have been deployed in public environments making evident the need of human-aware navigation capabilities. In this regard, the robotics community have made efforts to include proxemics or social conventions within the navigation approaches. Nevertheless, few works have tackled the problem of labelling humans as an interactive agent when blocking the robot motion trajectory. Current state of the art navigation planners will either propose an alternative path or freeze the motion until the path is free. We present the first prototype of a framework designed to enhance social competency of robots while navigating in indoor environments. The implementation is done using \textit{Navigation} and \textit{Object Detection} open-source software. Specifically, the Robot Operating System (ROS) navigation stack, and OpenCV with Caffe deep learning models and MobileNet Single Shot Detector (SSD), respectively.

\end{abstract}




%
% Keywords. The author(s) should pick words that accurately describe the work being
% presented. Separate the keywords with commas.

%
% A "teaser" image appears between the author and affiliation information and the body 
% of the document, and typically spans the page. 

\begin{teaserfigure}
\centering
  \includegraphics[width=0.7\textwidth]{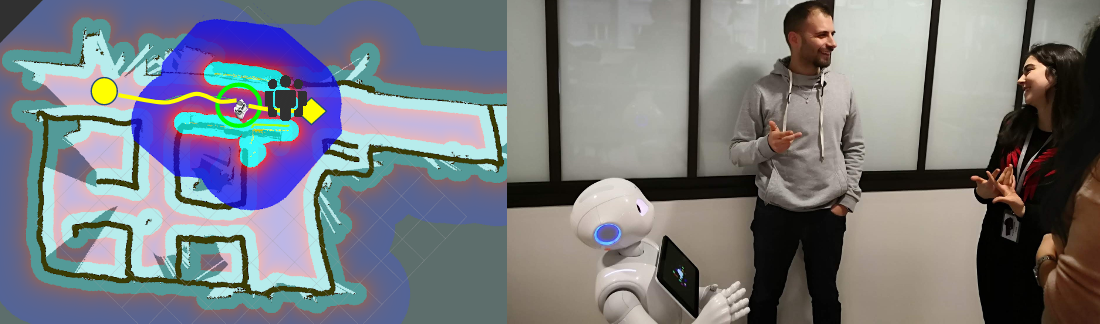}
  \caption{ROS map visualization and initial plan obstructed by people (Left). Real scenario (Right).}
  \Description{Pepper people perception enhanced by the MobileNet Single Shot Detector.}
  \label{fig:teaser}
\end{teaserfigure}

%
% This command processes the author and affiliation and title information and builds
% the first part of the formatted document.
\maketitle

\section{Introduction}

With the rise of social robots in our society, modelling interaction with people in real world scenarios is fundamental. Environments where robots are capable of navigating such as hospitals, hotels, restaurants, train stations or airports have been subject of research studies. In \cite{sasaki2017long}, a robot called Peacock was deployed 120 hours in a museum. The sessions were carried out two times per week for a lapse of ten months letting visitors move freely around the robot during autonomous navigation. The authors conclude that the approach performed well in uncongested scenarios, with limitations in crowded situations or narrow aisles. A similar example are the robots deployed in restaurants serving food in countries such as China or Japan, which typically use line followers for trajectory generation \cite{mishraa2018issues}. Recently, a ROS based robotic waiter was presented in \cite{cheong2016development}. The preliminary tests shown that the platform was able to dock at a target table and roll out the serving tray successfully.

Research studies in Human-Robot Interaction (HRI) have shown that robot behaviour has a deep impact on their perceived intelligence, especially in the case of service robots navigating in public spaces \cite{althaus2004navigation}. In this regard, several methods have been proposed for navigating on dynamic and uncertain environments. In summary, they can be classified in two groups: model-based and learning-based methods.

The major exponents of model-based methods rely on social psychology and cognitive sciences to generate human-like paths for robot navigation. One of the most relevant approaches is the Social Force Model (SFM) \cite{helbing1995social}, proposed to model pedestrian's behaviour whose motion is influenced by other pedestrians by means of repulsive forces. Many studies have implemented this method but also produced several variations \cite{zanlungo2011social}, later applied to real world environment navigation where robots can avoid or go along with people \cite{ferrer2013robot} \cite{ratsamee2013human}. However, these works show limitations such as the need for parameter calibration in different robots or the need of additional sensors for pedestrian tracking. Also, model-based methods are based on geometric relations, but it is still unclear if pedestrians always follow such models.

In contrast, learning-based methods use policies for defining human-like behaviours, which are usually learnt from human demonstrations by matching feature statistics about pedestrians. These methods apply machine learning techniques such as Inverse Reinforcement Learning (IRL) to model the factors that motivate people's actions instead of the actions themselves. An experimental comparison of features and learning algorithms based on IRL is presented in \cite{vasquez2014inverse}, where the authors conclude that is more effective to invest effort on designing features rather than on learning algorithms. More recent approaches include the use of deep reinforcement learning in order to set restrictions \cite{chen2017socially} (i.e. passing at the left side of the people) instead of learning the features that describe human paths.

Both groups of methods described above focus on the motion of the robot but there are situations where more complex social behaviors are needed. As pointed by \cite{trinh2015go}, polite navigation is an important requirement for social acceptance. According to this, an approach for management of deadlock situations at narrow passages is presented, where the robot lets the conflicting person pass and waits in a non-disturbing waiting position. A more recent work \cite{vega2018planning} proposes a navigation planning showing two scenarios: in the first one, the robot asks for collaboration to enter a room. In the second one, the robot asks for permission to navigate between two people which are talking. However, results presented are shown in simulation.

In the present work, we focus on a functional implementation and deployment of a similar scenario proposed in \cite{vega2018planning}. These kinds of situations challenge existing planners, yet resulting in a robot freeze while the path is blocked or re-generating a new alternative path, if possible. Instead, the most natural human behaviour is initiating a dialog to ask for permission to pass. In this regard, we propose the use of a high level situation assessment which is composed by a navigation module and a people perception module. This framework stops the navigation when a deadlock is generated by the situation previously described and triggers an episodic interaction before re-initiating the navigation once again. As a result, this work is the first of its kind in successfully incorporating human interaction as part of the navigation flow and deploy it on a robotic platform.

\section{Methodology}

The proposed methodology makes use of existing open source software to provide navigation and people perception capabilities. Because of this, it can be deployed on any robotic system able to run ROS and OpenCV version 3.3+. In this work, we have used the humanoid robot Pepper as deployment platform. Nevertheless, due to the limited computational power of Pepper, an external CPU (Intel Core i7-3770 CPU 3.40GHz x 8) with Ubuntu 16.04 LTS has been used to run the navigation and people perception modules.

\subsection{Robot Platform}

Pepper (version < 1.8) is an omnidirectional wheeled humanoid robot 1.21 m tall. With 17 joints and 20 degrees of freedom (DoF) kinematic configuration and edgeless design, it is suitable for social and safe Human-Robot Interaction \cite{pandey2018mass}. The platform is equipped with a large variety of sensors and actuators that ensure safe navigation; on its base, it incorporates 3 lasers pointing forward and to the sides, with a range of 60\textdegree. It also includes 2 sonars, which are pointing forward and backwards with a vertical and horizontal range of 60\textdegree, and 3 bumpers for object collision detection. Finally, the robot is equipped with an Atom E3845 Quad-core processor, 4 microphones, two 2D cameras and a depth camera in the head.

\subsection{Navigation}   

The navigation module is essentially composed by the ROS navigation stack including the packages Adaptive Monte Carlo Localization (\textit{amcl}) and Dynamic Window Approach (\textit{dwa\_local\_planner}).
Due to the low detection range of the Pepper lasers, a similar approach previously presented by \cite{perera2017setting} and \cite{suddrey2018enabling} has been used in order to convert the depth image into virtual laser data, using the ROS package \textit{depthimage\_to\_laserscan}. Then, the resulting map is generated using \textit{gmapping} (laser-based SLAM) and post-processed for improving its reliability. Similarly, a sketched map of the scenario was also used to improve the performance. The navigation uses a global planner with inflated obstacles and a local \textit{costmap} with
observations from the virtual laser data. In this way, Pepper will stay away from possible collisions benefiting the motion of the robot. Additionally, some necessary components\footnote{\url{https://bitbucket.org/account/user/pepper_qut/projects/PN}} to orchestrate the navigation have been incorporated \cite{suddrey2018enabling} and presented in the next section.

The planning task is carried out by two main components; the global planner, which is in charge of providing path trajectory from the initial location till the target goal, and the local planner, responsible for obstacle avoidance in a close range while keeping an optimal distance to the global path. In case of failure, two recovery behaviors are implemented allowing the \textit{costmap} to be cleaned and the robot rotate in place to find a new global path. However, we introduce a preliminary step where we expose the specific time the local planner could not find a valid plan in order to trigger the higher level situation assessment that allows the identification of deadlock situations generated by humans.

\subsection{People Perception}
\begin{figure}[h!]
  \centering
  \includegraphics[width=0.95\linewidth]{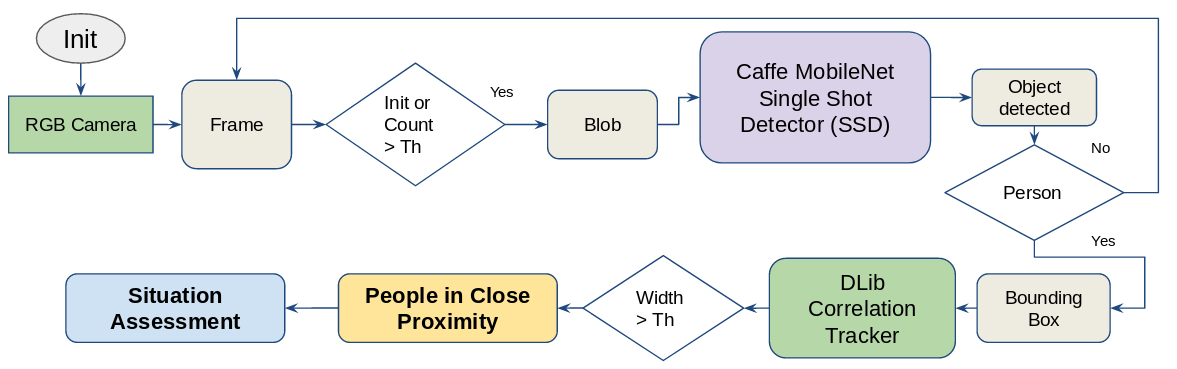}
  \caption{People Perception module working principle.}
  \Description{People Perception module working principle.}
  \label{fig:process-perception}
\end{figure}

Current state of the art algorithms based on Deep Learning such as YOLO \cite{redmon2016you} offer a fast object detection at a high computational cost. However, based on our computational restrictions, an accurate approach suitable for CPU processing was required. For this reason, the OpenCV \textit{dnn} module composed of a MobileNet-Single Shot Detector (SSD) \cite{howard2017mobilenets} trained in Caffe framework was chosen. This implementation uses an RGB image as input and it is able to detect up to 20 different classes, \textit{humans} among them, despite occlusions and from different points of view.
Once the person is detected, and with the aim to decrease the computational cost, a lightweight correlation tracker implemented on Dlib library \cite{danelljan2014accurate} is applied. The module publishes a message every time a bounding box is bigger than an empirically predefined threshold (80 pixels width) in order to filter targets located far away from the field of interaction (see figure \ref{fig:process-perception}).

\section{Situation Assessment Framework}

The high level situation assessment framework proposed makes use of the Navigation and People Perception modules to orchestrate the different transitions between them. As pointed earlier, detecting a failure in the local planner is key to trigger an episodic interaction with the human blocking, but also considering that this event can take place due to several reasons (i.e. a failure due to loss of expected rate in sensor data acquisition). In our case, we are not only interested to know if there is a failure due to an obstacle in front of the robot, but also if the robot has detected a person as a source. At the same time, these changes are used in the global plan in order to know if a newly generated trajectory is better than the original one.

The navigation cycle is initiated by the NAOqi application \textit{Navigation App} that loads the available destinations from a \textit{json} file into the \textit{Location server}. The same application is in charge of interfacing with ROS through \textit{ActionLib} client ROS package in order to generate the robot motion.

\begin{figure}[h!]
  \centering
  \includegraphics[width=0.85\linewidth]{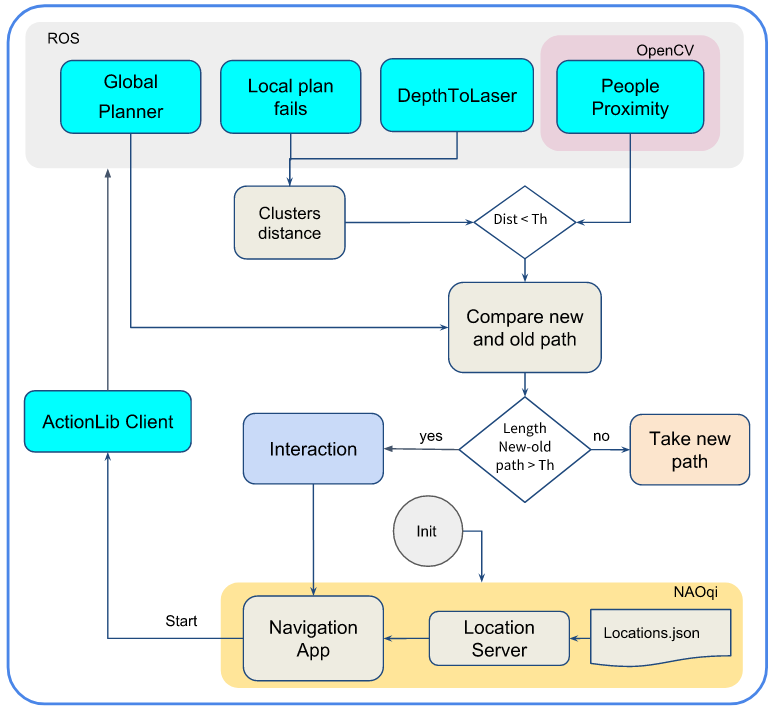}
  \caption{Situation Assessment workflow.}
  \Description{Situation Assessment workflow.}
\end{figure}

As in any complex system, the inputs of the sensors and the outputs from the Navigation and People Perception modules are obtained asynchronously, but sequentially evaluated in order to stop or start the navigation and the user interaction. When the local plan has failed, the data retrieved from the virtual laser is grouped into clusters using \textit{MeanShift} from the \textit{scikit-learn} library with a quantile value of 0.20 to estimate the bandwidth; if the mean of any of the clusters is smaller than a specified distance (<3 meters due to the depth sensor limitation range), the People Perception module engages during \textit{t} seconds in order to detect the human. In addition, if there is an existent plan generated by the global planner, a comparison with the initial trajectory is performed using the difference in path lengths. 

In consequence, if the newly generated path is shorter or equal than the original one, considering a threshold or \textit{level of consideration}, the robot still can take it avoiding the interaction. Such measurement becomes a very powerful tool in order to define constraints that respect specific social contexts and groups. For instance, if a physically impaired person is detected or the robot is deployed on a cluttered environment, this would allow to modulate the robot's behaviour or eventually exclude the interaction.

Once all the conditions described above are satisfied, the navigation is stopped and the robot starts the verbal interaction. In case the path remains blocked, the robot would wait 5 seconds and ask for permission a second time. Alternatively, the system can freeze till the path is no longer blocked.

\section{Results}

\begin{figure*}[h]
\centering
\subfloat[ROS Navigation plan with no obstacles][Absence of obstacles]{
\includegraphics[width=0.27\textwidth, height=3.3cm]{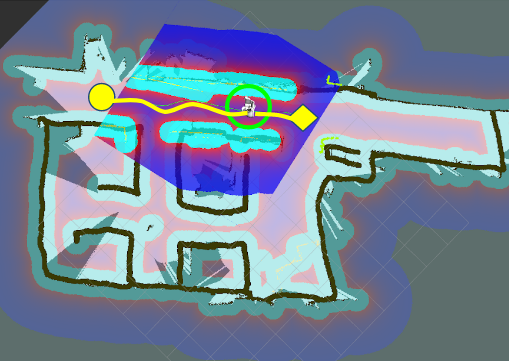}
\label{fig:subfig1}}
\subfloat[Interactive navigation with people blocking the path.][Navigation with people blocking the path]{
\includegraphics[width=0.27\textwidth, height=3.3cm]{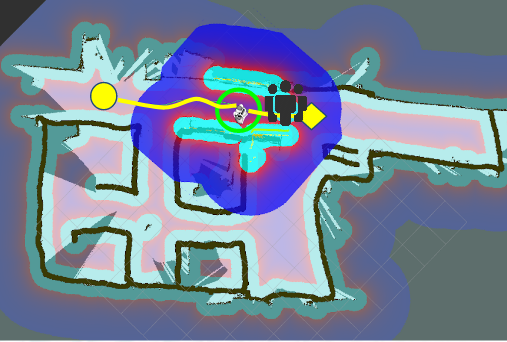}
\label{fig:subfig3}}
\subfloat[Real scenario people - giving access.][Real scenario - people blocking the path]{
\includegraphics[width=0.27\textwidth, height=3.3cm]{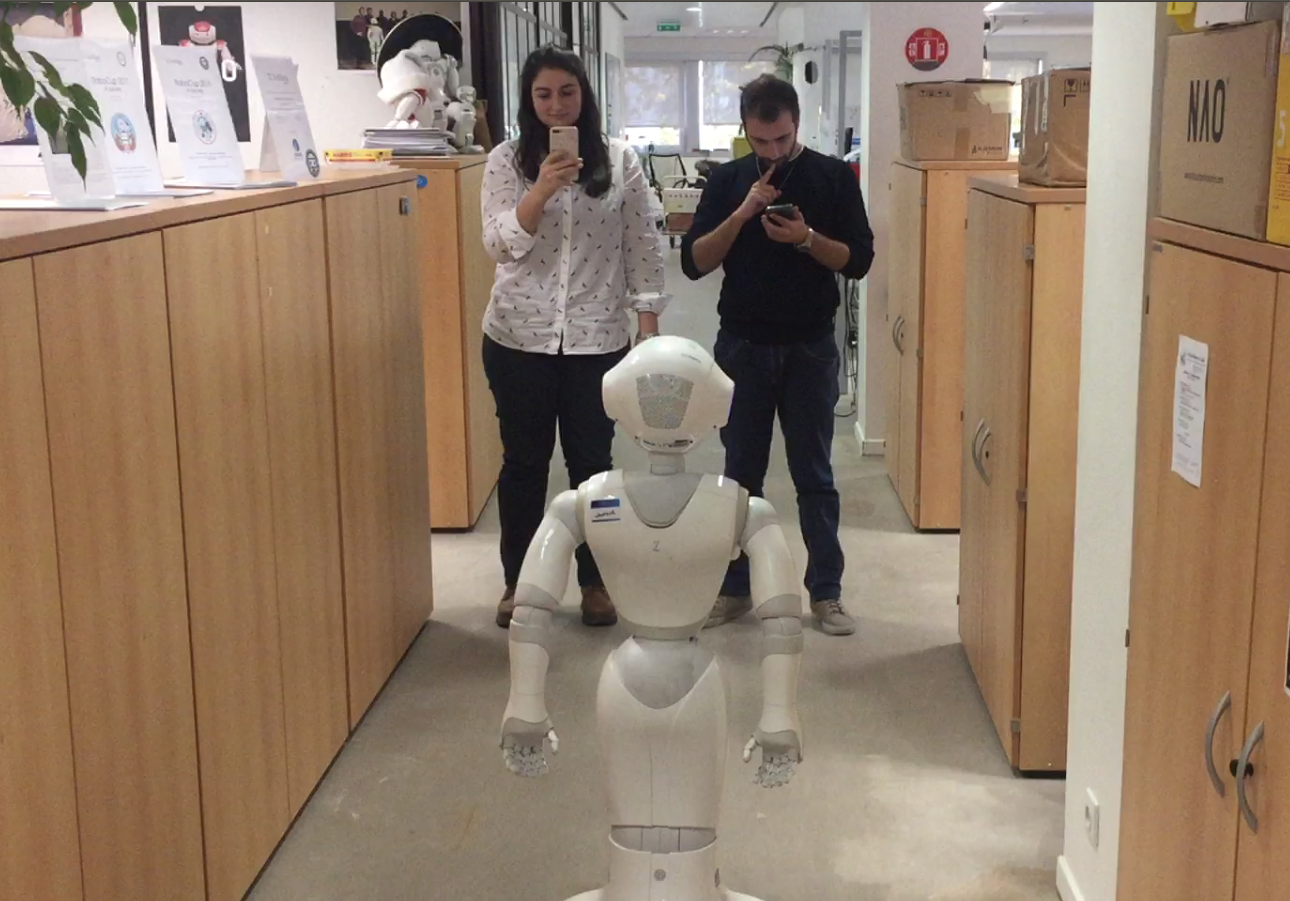}
\label{fig:subfig5}}
\qquad
\subfloat[ROS Navigation plan with obstacles][People blocking the path]{
\includegraphics[width=0.27\textwidth, height=3.3cm]{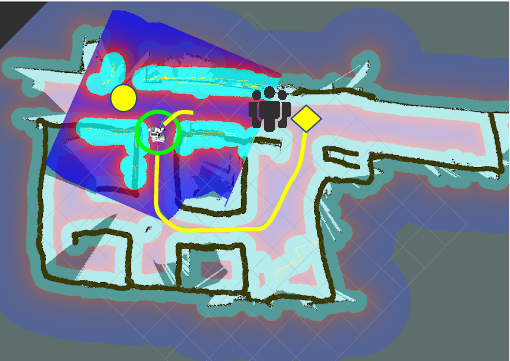}
\label{fig:subfig2}}
\subfloat[Interactive navigation after people free up the path.][Navigation after people free up the path]{
\includegraphics[width=0.27\textwidth, height=3.3cm]{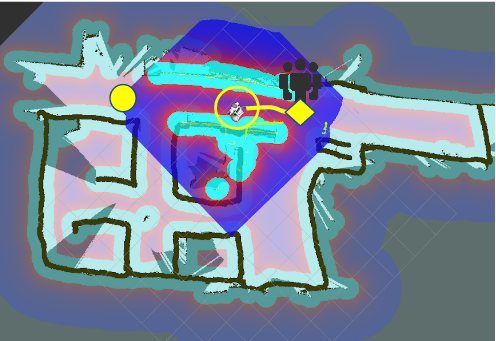}
\label{fig:subfig4}}
\subfloat[Real scenario people blocking the path][Real scenario - people interacting]{
\includegraphics[width=0.27\textwidth, height=3.3cm]{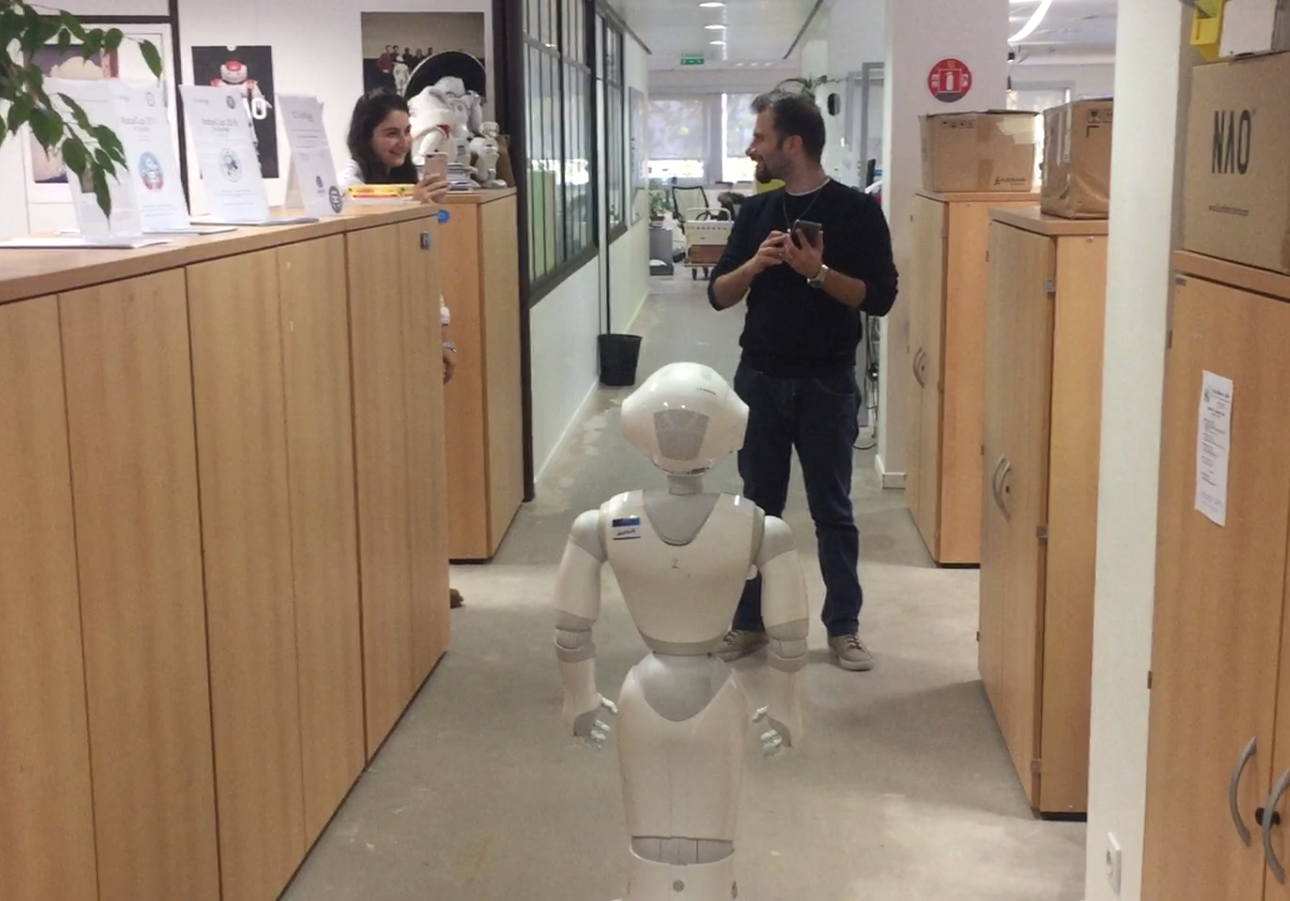}
\label{fig:subfig6}}
\caption{Generated plans with ROS navigation stack (a) and (d), and with the framework proposed (b), (c), (e) and (f)}
\label{fig:navigation}
\end{figure*}

Three cases where the robot is navigating in a hallway (figure \ref{fig:navigation}) are presented. The first one shows a free path and the other two people blocking it. Figure \ref{fig:subfig1} shows the robot taking the shortest path due to the absence of any type of obstacles. In contrast, when the path is blocked potentially due to a human (see figure \ref{fig:subfig2}), the robot changes its initial plan and takes a longer path to reach its goal. In figure \ref{fig:subfig3}, the robot detects that the path is blocked by humans and interacts with them in order to free it. Finally, it will take a new plan that is no longer than the initial one (figure \ref{fig:subfig4}).  

During our preliminary trials, it is worth mentioning that when the robot asked for permission to pass (figure \ref{fig:subfig5}), one of the people standing moved out while the other remain blocking the potential trajectory. Then, an unexpected human-robot collaboration took place: the person aware of the intention of the robot asked the other one to free the way (see figure \ref{fig:subfig6}).

\section{Conclusions and Future Work}

In the present work, we have successfully incorporated HRI as part of the navigation strategy by performing a situation assessment on a human blocking context. In addition, the system presented has been deployed on a humanoid robot in a real world scenario. However, several limitations have been identified during the preliminary tests. First of all, since the computation is not on-board, the segmentation of the network penalizes the performance of the people perception module. Secondly, the situation assessment module working principle is not embedded into the navigation planner; with an integration to the ROS local planner, a consolidated approach could be provided.

In terms of future work, further implementation needs to be done in order to include the use of social and individual situations to modulate the specified \textit{level of consideration}. In this way, we will be able to personalize the decision making for the interruption and asking permission to pass in a blocked path.

%
% The acknowledgments section is defined using the "acks" environment (and NOT an unnumbered section). This ensures
% the proper identification of the section in the article metadata, and the consistent spelling of the heading.
\begin{acks}
This work has received funding from the European Union's Horizon 2020 framework programme for research and innovation under the Industrial Leadership Agreement (ICT) No. 779942 (CROWDBOT).
\end{acks}
%
% The next two lines define the bibliography style to be used, and the bibliography file.
\bibliographystyle{ACM-Reference-Format}
\bibliography{prueba}

\end{document}